\documentclass{article}

\PassOptionsToPackage{numbers}{natbib}
\usepackage[preprint]{neurips_2026}

\makeatletter
\renewcommand{\@noticestring}{}
\makeatother

\usepackage[utf8]{inputenc}
\usepackage[T1]{fontenc}
\usepackage{hyperref}
\usepackage{url}
\usepackage{booktabs}
\usepackage{amsfonts}
\usepackage{amsmath}
\usepackage{nicefrac}
\usepackage{microtype}
\usepackage{xcolor}
\usepackage{graphicx}
\usepackage{subcaption}
\usepackage{algorithm}
\usepackage{algorithmic}
\usepackage{multirow}
\usepackage{pifont}

\usepackage{colortbl}
\usepackage{wrapfig}
\usepackage[most]{tcolorbox}
\usepackage{soul}
\usepackage{arydshln}
\sethlcolor{highlightblue}
\definecolor{rowgray}{HTML}{F5F5F5}
\definecolor{highlightblue}{HTML}{E3EDF7}

\definecolor{cmarkgreen}{RGB}{34, 139, 34}
\definecolor{xmarkred}{RGB}{200, 50, 50}
\definecolor{rowgray}{RGB}{234, 240, 245}
\definecolor{highlightblue}{RGB}{200, 240, 255}
\definecolor{rank1}{RGB}{230, 140, 140}
\definecolor{rank2}{RGB}{240, 175, 175}
\definecolor{rank3}{RGB}{250, 210, 210}
\definecolor{bluerank1}{RGB}{200, 220, 245}
\definecolor{bluerank2}{RGB}{220, 235, 250}
\definecolor{bluerank3}{RGB}{240, 245, 255}

\graphicspath{{figures/}}

\title{LiveBrowseComp: Are Search Agents Searching, or Just Verifying What They Already Know?}

\author{%
  HuiMing Fan$^{1,*}$ \quad
  Xiao Wang$^{2,*,\dagger}$ \quad
  Zheng Chu$^{1}$ \\
  Qianyu Wang$^{1}$ \quad
  Zhuoyao Wang$^{1}$ \quad
  Ming Liu$^{1,\dagger}$ \quad
  Bing Qin$^{1}$ \quad
  XingYu$^{2}$ \\
  $^{1}$Harbin Institute of Technology \quad
  $^{2}$Xiaohongshu
}

\begin{document}

\maketitle

\begingroup
\renewcommand{\thefootnote}{\fnsymbol{footnote}}
\footnotetext[1]{Equal contribution. \quad $^\dagger$Corresponding authors.}
\footnotetext[2]{\texttt{hmfan@ir.hit.edu.cn}, \texttt{wangxiao14@xiaohongshu.com}, \texttt{mliu@ir.hit.edu.cn}.}
\endgroup

\vspace{-3mm}
\begin{abstract}
Are LLM-based search agents genuinely searching, or using the web to verify what they already know? 
We study this question on BrowseComp with three diagnostics. 
Our analysis reveals Intrinsic Knowledge Dependence (IKD): even with tool access, agents often rely on intrinsic knowledge---information encoded in the model before retrieval---rather than on external evidence.
Agents answer up to 44.5\% of BrowseComp questions without tools, generate more than half of their search queries from internally produced hypotheses rather than retrieved leads, and perform worse than closed-book baselines when answer-supporting evidence is removed.
These results suggest that static search benchmarks can reward memory-backed verification rather than evidence-driven discovery, conflating what agents already know with what they can find.
We then introduce LiveBrowseComp, a deep-search benchmark designed to evaluate agents beyond intrinsic coverage.
It contains 335 human-authored questions whose answers depend on facts published within the 90 days preceding benchmark construction, drawn from six updated sources and filtered to exclude globally salient events.
On LiveBrowseComp, all evaluated agents fall below 2\% closed-book accuracy, search-augmented scores drop by 25–40 points relative to BrowseComp, and prior model rankings no longer reliably predict performance. 
LiveBrowseComp is available at \url{https://huggingface.co/datasets/Forival/LiveBrowseComp}.
\end{abstract}

\vspace{-3mm}
\section{Introduction}
\label{sec:intro}

Large language models (LLMs)~\cite{achiam2023gpt,comanici2025gemini,zeng2025glm} are increasingly deployed as autonomous agents rather than mere text generators. Search agents~\cite{jin2025search,team2025tongyi,chu2026redsearcher} are a central example: they browse the web, integrate evidence across sources, and answer complex information needs. Systems such as OpenAI Deep Research~\cite{openai_deep_research_2026} and Gemini Deep Research~\cite{google_gemini_deep_research_2025} show how rapidly this direction is being deployed. Evaluation has evolved in parallel, from single-turn QA (TriviaQA~\cite{joshi2017triviaqa}, NaturalQuestions~\cite{kwiatkowski2019natural}) and multi-step reasoning (HotpotQA~\cite{yang2018hotpotqa}) to agentic web-search benchmarks such as BrowseComp~\cite{browsecomp2025} and DeepSearchQA~\cite{gupta2026deepsearchqa}. On BrowseComp, leading models~\cite{openai_gpt55_2026,anthropic_opus46_2026,minimax_m2_5_2026,kimi_k2_6_2026} have posted increasingly high scores. Yet a fundamental question arises: \textbf{are these scores evidence that agents are genuinely searching, or are agents merely using the web to verify what they already know?}

To answer this question, we design a set of diagnostic experiments that progressively remove or perturb the role of retrieved evidence. The diagnostics ask three simple questions. First, if search benchmarks truly require search, how well can agents answer them with all tools removed? Second, if agents use tools for discovery, what happens when the search environment is intact but all answer-supporting evidence is removed? Third, during multi-step browsing, do agents actually build new hypotheses from retrieved evidence, or do they continue querying entities already produced by their own internal knowledge? Together, these experiments isolate whether tool use is driving the answer, or whether the web is being used primarily as a verification interface for parametric knowledge.

\begin{figure}[t]
  \centering
  \includegraphics[width=\linewidth]{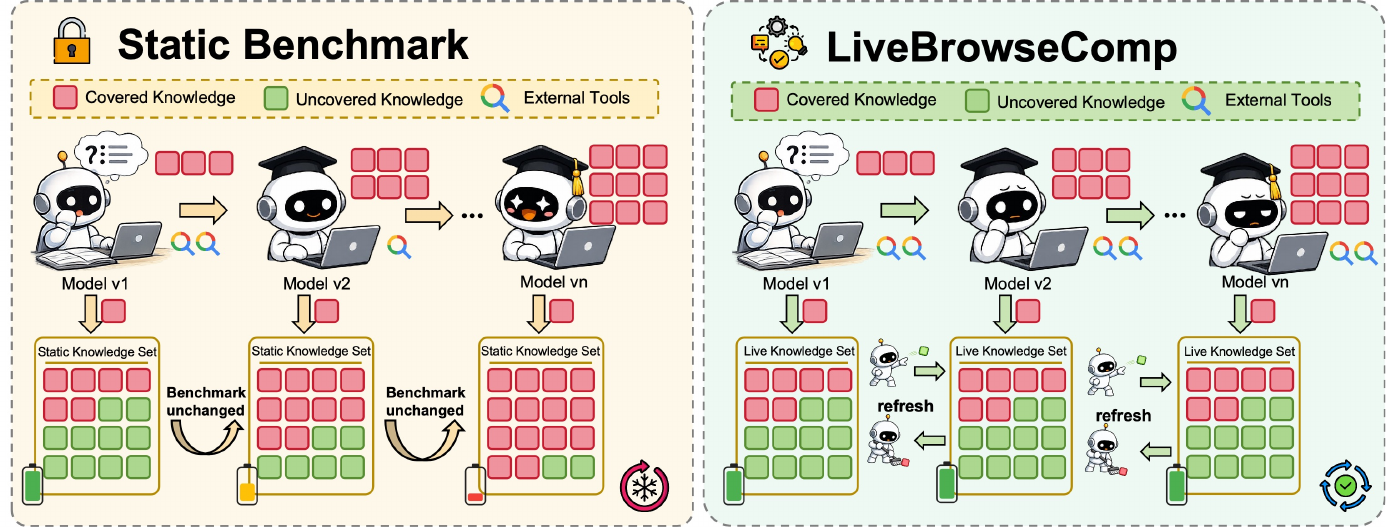}
  \caption{Overview of LiveBrowseComp. As models iterate, the knowledge required by a static benchmark is gradually absorbed into their parameters, so the effective difficulty of its questions collapses over time. By being constructed from up-to-date knowledge, LiveBrowseComp can effectively mitigate this erosion.}
  \label{fig:overview}
\end{figure}

The diagnostics reveal a simple but troubling pattern. Many benchmark questions are already covered by agents' intrinsic knowledge---parametric knowledge available without retrieval: \textbf{with all search tools removed, closed-book pass@4 reaches up to 44.5\%}, and every evaluated model obtains non-trivial scores across existing benchmarks. More importantly, \textbf{search becomes harmful when it can no longer verify this intrinsic knowledge}. In an evidence-blocking setting, where the search interface remains available but all answer-supporting documents are removed, every model performs worse than its closed-book baseline: MiniMax M2.5~\cite{minimax_m2_5_2026} drops from 44.5\% to 8.0\%, and Kimi-K2.6~\cite{kimi_k2_6_2026} from 25.5\% to 2.3\%. Trajectory analysis explains why: more than half of agents' queries are seeded by information that first appears in the model's own reasoning rather than in retrieved documents; after failed searches, agents often only rephrase the previous query; and even when useful evidence is retrieved, they frequently fail to use it.

We call this failure mode \emph{Intrinsic Knowledge Dependence} (IKD). Under IKD, agents appear effective on static benchmarks because they can guess from memory and use search for confirmation; but when the needed fact lies outside their knowledge boundary, the search loop loses its anchor and collapses. This is not merely data contamination: even uncontaminated questions can be solved through broad parametric world knowledge. As models become more knowledgeable, fixed benchmarks increasingly reward memory-backed verification rather than genuine search, conflating what a model already knows with how well it can discover what it does not know.

To evaluate search capability beyond this shortcut, we introduce \textbf{LiveBrowseComp}, a deep-search benchmark designed to sit outside models' current knowledge boundary. It contains 335 human-authored questions, each depending on facts published within the 90 days preceding benchmark construction and unanswerable from earlier information alone. Questions are seeded from six continuously updated sources---GDELT~\cite{gdelt2026}, TMDB~\cite{tmdb2026}, RAWG~\cite{rawg2026}, CVE/NVD~\cite{cve_nvd2026}, SportsDB~\cite{sportsdb2026}, and USGS~\cite{usgs2026}---and filtered to exclude globally salient events, retaining obscure but publicly verifiable facts. Each question is independently validated by human verifiers using only web search to ensure solvability and uniqueness. Archived benchmark snapshots are preserved for reproducibility.

LiveBrowseComp exposes the gap hidden by static benchmarks. \textbf{Every evaluated model falls below 2\% closed-book accuracy}, showing that the temporal and long-tail constraints largely neutralize intrinsic knowledge. Once this memory backstop is removed, \textbf{search-augmented scores drop by roughly 25--40 points relative to BrowseComp}, and static-benchmark rankings no longer reliably predict performance. Human searchers, however, require comparable effort on LiveBrowseComp and BrowseComp, indicating that the drop is not caused by intrinsically harder questions. LiveBrowseComp therefore isolates the failure mode: agents struggle not because the tasks are unsolvable, but because memory-backed verification no longer works. It shifts evaluation from confirming what agents already know to discovering what they do not.

\section{Pilot Study}
\label{sec:pilot}

Frontier search agents have achieved strong results on challenging browsing benchmarks, but the source of this success remains unclear. An agent may discover an answer by following evidence obtained through search, or it may first generate a plausible hypothesis from intrinsic knowledge and then use search primarily to confirm it. 

We conduct the pilot study on four challenging agentic benchmarks: BrowseComp~\cite{browsecomp2025}, BrowseComp-ZH~\cite{zhou2025browsecomp}, HLE~\cite{phan2025humanity}, and GAIA~\cite{mialon2023gaia}. These benchmarks cover complementary evaluation settings, including long-horizon web browsing, multilingual browsing, expert-level knowledge reasoning, and general tool-augmented problem solving. We evaluate recent frontier agentic models from both open-source and closed-source families~\cite{liu2025deepseek,zeng2026glm,minimax_m2_5_2026,kimi_k2_6_2026,deepseek_v4_2026,seed2026seed2}, since these systems represent the strongest current search-agent capabilities and are also most likely to possess broad intrinsic knowledge. To separate these two modes, we conduct three diagnostics:


\begingroup
\setlength{\leftmargini}{1.6em}
\setlength{\labelsep}{0.4em}
\renewcommand{\labelenumi}{\textbf{Q\arabic{enumi}.}}
\begin{enumerate}
    \setlength{\itemsep}{1pt}
    \setlength{\parskip}{0pt}
    \setlength{\parsep}{0pt}
    \item \textit{Closed-book coverage} estimates how much benchmark-relevant knowledge agents can already produce without retrieval;
    \item \textit{Evidence-blocked search} tests whether tool use remains beneficial when answer-supporting documents are removed from the retrieval environment;
    \item \textit{Trajectory grounding} examines whether subsequent queries are grounded in retrieved evidence or seeded by hypotheses generated by the model itself.
\end{enumerate}
\endgroup

Together, these diagnostics test whether search functions as a discovery mechanism or mainly as a verification interface for intrinsic knowledge. For tool-use experiments, we use a unified search-agent scaffold~\cite{chu2026redsearcher} with a shared interaction protocol, sampling budget, context limit, and answer format across models. Closed-book experiments use the same sampling and answer-format constraints but remove all tools. For evidence-blocking and trajectory analysis, we use BrowseComp-Plus~\cite{chen2025browsecomp}, which provides annotated evidence, gold, irrelevant, and hard-negative documents for each question. We construct a dense retrieval index over this document library using Qwen3-8B-Embedding~\cite{yang2025qwen3} and expose it through the same search interface across models. In the blocked condition, evidence and gold documents are removed from the index, leaving only irrelevant and hard-negative documents. This controlled setting lets us manipulate evidence availability and analyze query provenance while reducing variance from live-web ranking, crawling failures, and page availability.

\subsection{Answering without Tools: Measuring Knowledge Coverage}
\label{sec:pilot:closedbook}

We first ask how much benchmark performance is already available before search begins.
Closed-book answering does not prove memorization, but it provides a conservative proxy for intrinsic knowledge coverage: if an agent answers correctly with all tools removed, the success cannot be attributed to retrieval.
We therefore disable all search tools and require each model to answer using only its parametric knowledge across four benchmarks.
Implementation details are provided in Appendix~\ref{app:closedbook}.

\begin{figure}[t]
  \centering
  \includegraphics[width=\linewidth]{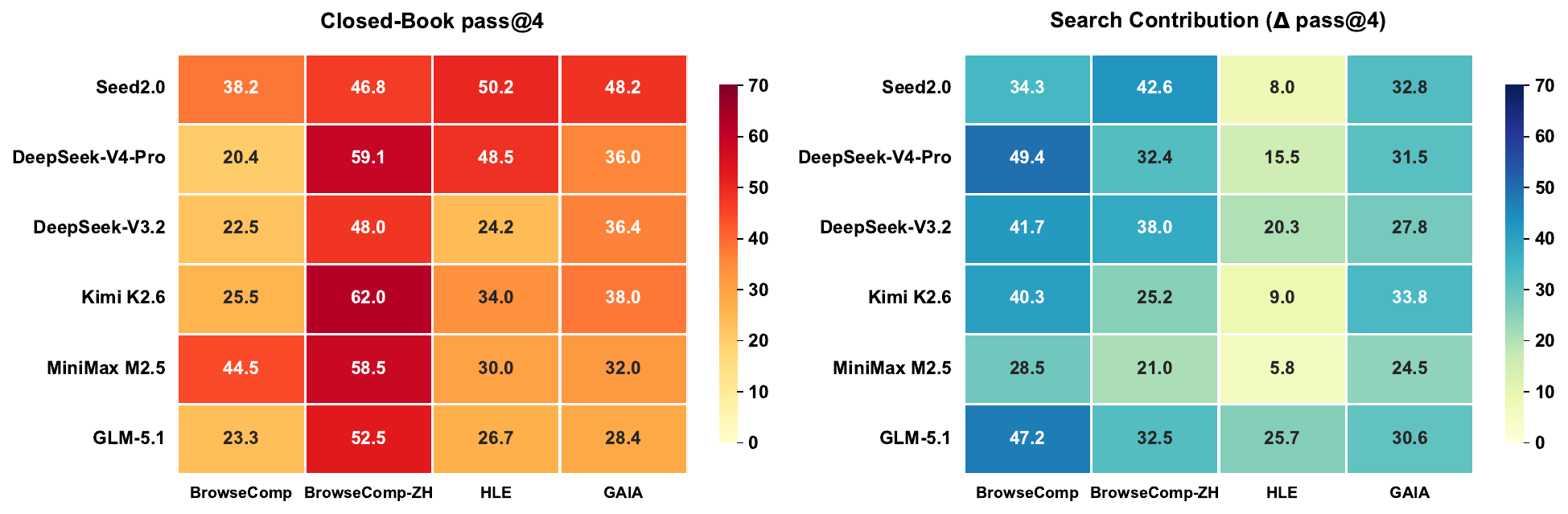}
    \caption{Closed-book performance and tool-use gains on static search benchmarks. 
    Left: pass@4 without access to tools. 
    Right: the absolute gain from tools, computed as pass@4 with tools minus pass@4 without tools. 
    Closed-book performance is already substantial, and the models that benefit most from tools are not necessarily those with the strongest closed-book coverage.}
  \label{fig:closedbook}
\end{figure}

Figure~\ref{fig:closedbook} shows that closed-book performance accounts for a substantial fraction of benchmark success.
Across all 24 model–benchmark pairs, pass@4 ranges from 20.4 to 62.0, averaging 38.9.
Several results are especially striking: Kimi K2.6 reaches 62.0 on BrowseComp-ZH, MiniMax M2.5 reaches 44.5 on BrowseComp, and Seed 2.0 reaches 50.2 on HLE, all without retrieval.
\textbf{Thus, a substantial fraction of performance on existing ``search'' benchmarks is already available before any search is performed.}

Tool access further improves performance, but the pattern of improvement does not simply mirror closed-book strength.
For example, MiniMax M2.5 obtains the highest closed-book score on BrowseComp, yet its search contribution is relatively modest at 28.5 points; in contrast, DeepSeek-V4-Pro starts from a much lower closed-book score of 20.4 but gains 49.4 points from search.
Similarly, models with strong closed-book coverage on BrowseComp-ZH, such as Kimi K2.6 and MiniMax M2.5, do not receive the largest tool-use gains.
On HLE, tool-induced gains are generally limited across several models, with MiniMax M2.5, Seed 2.0, and Kimi K2.6 improving by only 5.8, 8.0, and 9.0 points, respectively.
These mismatches indicate that final benchmark scores conflate two different capabilities: knowing plausible answers before search begins and discovering answers through retrieval.
Closed-book coverage therefore establishes the first condition for intrinsic knowledge dependence: many benchmark questions can be answered, at least by some frontier agents, before search is used at all.

\vspace{-2mm}
\subsection{Searching with Tools: Blocking Answer-Supporting Evidence}
\label{sec:pilot:drop}

\begin{wraptable}{r}{6.2cm}
\centering
\small
\vspace{-4mm}
\caption{Evidence-blocked search hurts performance relative to closed-book answering (pass@4).}
\label{tab:drop100}
\setlength{\tabcolsep}{4pt}
\begin{tabular}{lcc}
\toprule
\textbf{Model} & \textbf{Closed} & \textbf{Blocked} \\
\midrule
GLM 5.0~\cite{zeng2026glm}              & 21.3 & 7.4 {\scriptsize $\downarrow$13.9} \\
GLM 5.1~\cite{zeng2026glm}              & 23.3 & 9.4 {\scriptsize $\downarrow$13.9} \\
MiniMax M2.5~\cite{minimax_m2_5_2026}   & 44.5 & 8.0 {\scriptsize $\downarrow$36.5} \\
Kimi-K2.5~\cite{kimi_k2_6_2026}         & 19.7 & 2.8 {\scriptsize $\downarrow$16.9} \\
Kimi-K2.6~\cite{kimi_k2_6_2026}         & 25.5 & 2.3 {\scriptsize $\downarrow$23.2} \\
DeepSeek-V4-Pro~\cite{deepseek_v4_2026} & 22.5 & 7.0 {\scriptsize $\downarrow$15.5} \\
\midrule
\textbf{Avg.}                           & \textbf{26.1} & \textbf{6.2 {\scriptsize $\downarrow$19.9}} \\
\bottomrule
\end{tabular}
\end{wraptable}

Closed-book accuracy shows that agents can often produce correct answers before retrieval.
We next ask whether search remains useful when the environment can no longer provide confirming evidence.
Using BrowseComp-Plus, we remove all evidence and gold documents from the dense retrieval index, leaving only irrelevant and hard-negative documents.
Agents can still issue queries normally, but the retrieved results no longer contain documents that support the correct answer.
Implementation details are provided in Appendix~\ref{app:config:bcplus}.

Table~\ref{tab:drop100} shows a consistent reversal: evidence-blocked search underperforms closed-book answering for every model.
Average pass@4 drops from 26.1 in the closed-book setting to 6.2 when answer-supporting evidence is blocked, and all blocked scores remain below 10.
The largest drops occur for models with substantial closed-book accuracy: MiniMax M2.5 falls from 44.5 to 8.0, and Kimi-K2.6 from 25.5 to 2.3.
\textbf{Across all evaluated models, searching with answer-supporting evidence removed performs worse than not searching at all.}

This reversal suggests that agents do not reliably treat retrieval as an evidence-discovery process.
A robust search agent should discount uninformative results and preserve a plausible answer when search fails to find support.
Instead, non-supporting retrieval consistently degrades performance, indicating that the search loop can pull agents away from correct intrinsic answers and into hard-negative trajectories.
In this setting, search behaves less like a mechanism for discovering evidence and more like a confirmation channel for internally generated hypotheses.

\begin{figure}[t]
  \centering
  \includegraphics[width=\linewidth]{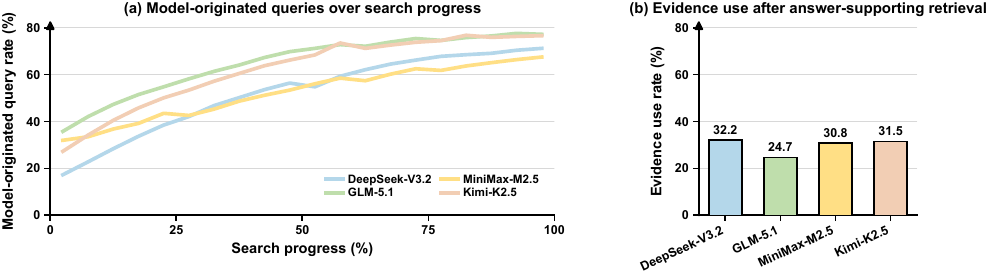}
  \caption{
    Search behavior on BrowseComp-Plus.
    Left: model-originated query rate over browsing progress.
    Right: evidence use rate after answer-supporting retrieval.
    Agents increasingly search from their own hypotheses and often fail to use retrieved evidence.
  }
  \label{fig:strategy}
\end{figure}

\subsection{Search Strategy Analysis}
\label{sec:pilot:trajectory}

We next inspect search trajectories to explain why evidence-blocked search can perform worse than closed-book answering.
For each query, we trace where its key information first appears.
If the information first appears in the model's own reasoning, we call the query a \textit{model-originated query}; if it first appears in retrieved results, we call it a \textit{retrieval-originated query}.
We also measure whether the model uses answer-supporting evidence after it has been retrieved: an answer-supporting retrieval is counted as used if the evidence appears in the model's reasoning or final answer within the next three rounds.

Figure~\ref{fig:strategy} shows that search is largely model-led.
For every model, more than half of all queries are model-originated, and this fraction increases as browsing proceeds, exceeding 60\% in later rounds.
In other words, agents do not primarily extend their search from retrieved leads; instead, they continue generating new search directions from their own hypotheses.

\textbf{Even when answer-supporting evidence is retrieved, agents often fail to use it.}
The evidence-use rate remains below one-third across all evaluated models: 32.2\% for DeepSeek v3.2, 24.7\% for GLM-5.1, 30.8\% for MiniMax M2.5, and 31.5\% for Kimi-K2.5.
Thus, the failure is not only retrieval-side: agents may retrieve the right evidence but still fail to let it redirect the search or determine the final answer.

These trajectory patterns explain the blocked-search collapse in Section~\ref{sec:pilot:drop}.
Agents mainly search from internally generated hypotheses and use retrieval to seek support for them.
When support is absent, they do not reliably fall back or pivot to retrieved alternatives; when support is present, they often fail to incorporate it.
The resulting loop is model-led rather than evidence-led.
\subsection{From Diagnosis to Benchmark Design}
\label{sec:pilot:summary}

Together, the three diagnostics identify a common failure mode that we call \emph{Intrinsic Knowledge Dependence} (IKD): agents use parametric knowledge to generate hypotheses and use retrieval mainly to confirm them. 
\textbf{The key problem is that current search benchmarks can reward knowing what to search for, rather than the ability to discover what is not already known.} 
As model knowledge expands, fixed question pools increasingly mix two factors that should be evaluated separately: intrinsic knowledge coverage and evidence-driven search. 
This creates a benchmark-design requirement: evaluation must place agents beyond their current knowledge boundary, where internally generated guesses are unlikely to suffice. 
The next section introduces \textsc{LiveBrowseComp}, a benchmark built from recent, long-tail facts whether agents can search when they do not already know what to verify.

\section{LiveBrowseComp: A Deep Search Benchmark Designed to Suppress IKD}
\label{sec:benchmark}

\begin{figure}[t]
\centering
\includegraphics[width=\textwidth]{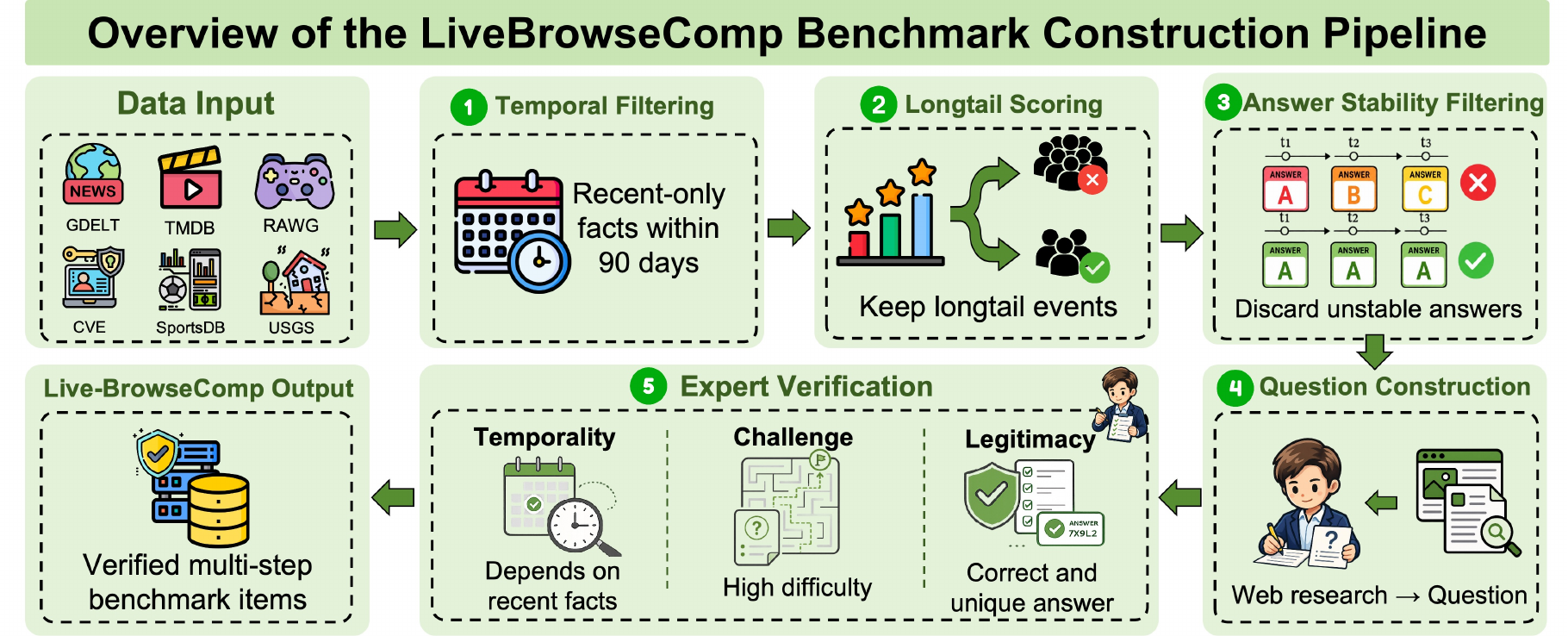}
\caption{The LiveBrowseComp construction pipeline, from seed sources through temporal and long-tail filtering, human annotation, and multi-dimensional verification to the final question bank.}
\label{fig:pipeline}
\end{figure}

The pilot study shows that search-agent evaluation must separate knowing plausible answers from discovering unknown information through evidence. 
We introduce LiveBrowseComp, a deep-search benchmark designed to sit outside models' current intrinsic knowledge coverage. 
Its questions rely on facts from the most recent 90 days and exclude globally salient events. 
They are also deliberately challenging: each question requires multi-step search and synthesis, targeting cases that ordinary users cannot solve within roughly 30 minutes. 
The aim is to remove the memory-backed verification shortcut, not to increase difficulty through obscurity alone. 
Figure~\ref{fig:pipeline} summarizes the construction pipeline, from time-bounded seed collection to filtering, question writing, and verification.

\subsection{Seed Collection and Filtering}
\label{sec:benchmark:criteria}

To place evaluation beyond the reach of intrinsic knowledge, seed selection enforces two constraints: \textit{recency}, which places queried facts beyond the likely training-data horizon, and \textit{obscurity}, which limits their exposure through widespread reporting; together, these constraints reduce the likelihood that the facts are encoded in the model's parametric memory.

We use six structured, continuously updated factual sources: GDELT~\cite {gdelt2026} for global news events, TMDB~\cite {tmdb2026} for film and television, RAWG~\cite {rawg2026} for video games, CVE/NVD~\cite {cve_nvd2026} for cybersecurity disclosures, SportsDB~\cite {sportsdb2026} for sports matches, and USGS~\cite {usgs2026} for earthquake records. Their public APIs provide timestamped records for precise temporal control, while their domain diversity mitigates the effect of any single-domain model advantage. We then extract candidate events from each source and apply three filters.

\paragraph{Stage 1: Temporal filtering.}
Intrinsic knowledge is accumulated during training. To push answers beyond this coverage, we discard any event whose core facts could be determined from information older than 90 days. The 90-day window comfortably exceeds typical data-collection lags in current training pipelines.

\paragraph{Stage 2: Long-tail filtering.}
Temporal recency does not guarantee that a fact falls outside intrinsic knowledge. Globally salient events can be absorbed into model parameters within days through post-training updates and reinforcement learning. To further reduce this overlap, we score each candidate on source-specific obscurity metrics such as audience reach, popularity counts, and mainstream coverage volume, and retain only events above a per-source long-tail threshold. Detailed criteria are provided in Appendix~\ref{app:datasources}.

\paragraph{Stage 3: Answer stability filtering.}
To ensure that each question has a single correct answer throughout the benchmark's lifespan, we remove candidates whose answers may change within the 90-day window. Cumulative box-office revenue, live standings, and chart rankings, for example, update progressively and do not settle at a fixed value. Only events with stable, uniquely determined answers proceed to question construction.

\subsection{Question Construction and Verification}
\label{sec:benchmark:verification}

We recruit professional annotators with undergraduate degrees or higher, strong English proficiency, and prior experience in data annotation. As screening and training, each annotator independently solves ten BrowseComp questions using only web search, must spend at least two hours before giving up, and must solve at least two out of ten correctly. This calibration ensures that every annotator internalizes the target difficulty and question type before contributing.

\paragraph{Stage 4: Question construction.}
After screening, annotators receive filtered seed events and independently conduct web research to craft questions. This involves: (1)~formulating a multi-step, multi-source reasoning question whose answer cannot be found in the first three pages of search engine results for the question text or any trivial reformulation of it; (2)~drafting a reference answer that is verifiable from definitive sources, confirming that the question admits exactly one short-string answer with no ambiguity; and (3)~anchoring at least one clue in a fact produced within the past 90 days, ensuring the question is unanswerable without this temporally recent information.

During construction, annotators document every web page they visit and assemble a complete evidence chain linking the question to the answer. This evidence chain serves as the primary input for Stage~5.

\paragraph{Stage 5: Peer Review.}
After construction, each question undergoes independent review by a separate verification team that was not involved in Stage~4. The review proceeds through three concurrent checks, designed to detect and eliminate questions that fail to meet the design criteria.

\textit{(a)~Correctness and uniqueness.} Verifiers trace the annotator’s evidence chain, visit each cited page, and confirm that the reference answer genuinely satisfies every constraint. To verify uniqueness, we employ multiple LLMs to generate a broad pool of candidate answers. Verifiers then manually check whether any candidate other than the reference answer satisfies all constraints (detailed protocol in Appendix~\ref{app:annotation}). Questions with broken evidence chains, logical gaps, or more than one valid answer are removed.

\textit{(b)~Difficulty calibration.} Independent annotators who were not involved in Stage~4 or check (a) attempt to solve each question using only web search. Each question is assigned to three annotators; if any annotator solves it within 30 minutes, the question lacks sufficient difficulty and is excluded.

\textit{(c)~Temporality verification.} Verifiers examine the evidence chain and identify every page whose content originates from within the past 90 days. For each such page, verifiers search for substitute evidence published before the 90-day window. If pre-window substitutes can be found for all temporally recent pages, the question is deemed not to genuinely depend on recent information and is discarded.

Across all three checks, each question is independently assessed by three verifiers who are mutually anonymous, and their results are further cross-checked by a fourth independent verifier to ensure annotation correctness. Detailed annotation workflow and compensation are provided in Appendix~\ref{app:annotation}.

\subsection{Dataset Examples}
\label{sec:benchmark:examples}

Compared to BrowseComp, LiveBrowseComp questions embed temporal constraints that tie the solution path to recently produced facts. For example, a typical BrowseComp question asks solvers to identify an entity from a set of static, time-independent clues (e.g., \textit{``Please tell me the name of the learning institution that\ldots{} in 2002, it held a three-day event\ldots{} in 2003, it held its graduation ceremony\ldots{}''}). All clues refer to fixed historical facts; a model with sufficient parametric knowledge can answer without any search.

In contrast, LiveBrowseComp questions interleave such multi-step reasoning clues with at least one temporally anchored constraint that cannot be resolved from pre-existing knowledge. Table~\ref{tab:examples} shows two examples. These temporal anchors force agents to retrieve recent evidence rather than rely on parametric memory, while the remaining clues preserve the multi-hop reasoning depth of BrowseComp.

\begin{table}[b]
\vspace{-6mm}
\centering
\footnotesize
\caption{Example questions from LiveBrowseComp. The temporally constrained clue in each question is highlighted in blue.}
\label{tab:examples}
\begin{tabular}{p{0.48\textwidth} p{0.48\textwidth}}
\toprule
\textbf{Example 1} & \textbf{Example 2} \\
\midrule
\hl{In 2026, a CVE vulnerability has been publicly disclosed. It is a flaw affecting server interfaces, targeting files with specific filename prefixes.} A malicious link can be constructed, and clicking the link will execute the corresponding code automatically. In addition, a vulnerability with a CVSS 3.x score 0.7 points higher than that of the server interface vulnerability was once found in an email marketing management tool. Attackers can launch attacks via two PHP files with highly similar names, and this vulnerability is of the same type as the server interface vulnerability. Could you please tell me in which year the vulnerability of this email marketing management tool occurred?(Answer: {\color{cmarkgreen}2020}) &
The name of the product is inspired by a term related to the concept of \"labor\". That precise metric served as the exclusive subject of a technical report released approximately \hl{three years earlier than the total solar eclipse that occurred in North America,} authored by an agency whose acronym doubles as an English interrogative pronoun and denotes the globe's primary intergovernmental health body. The tool's developer is a vendor based in Canada, with a name incorporating a term for a bird that builds a home. Based on these interconnected clues, what is the exact, case-sensitive name of the software product affected by the vulnerability? (Answer: {\color{cmarkgreen}WorkTime}) \\
\bottomrule
\end{tabular}
\end{table}

\subsection{Dataset Statistics}
\label{sec:benchmark:stats}

The current version of LiveBrowseComp contains 335 questions spanning eight topical categories. We aim to cover a broad range of topics to test model capability, with the number of questions per category corresponding to the distribution of available events after the filtering pipeline described in Section~\ref{sec:benchmark:criteria}.
\begin{figure}[t]
\centering
\includegraphics[width=\columnwidth]{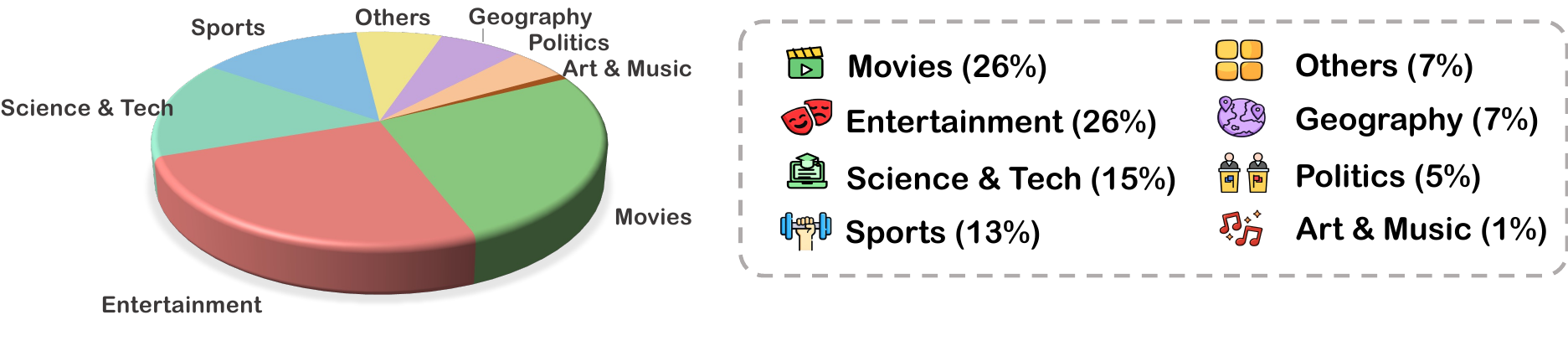}
\caption{Category distribution of LiveBrowseComp questions.}
\label{fig:domain_dist}
\end{figure}

\subsection{Human Performance on LiveBrowseComp}
\label{sec:benchmark:human}

\begin{figure}[h!]
\centering
\includegraphics[width=\textwidth]{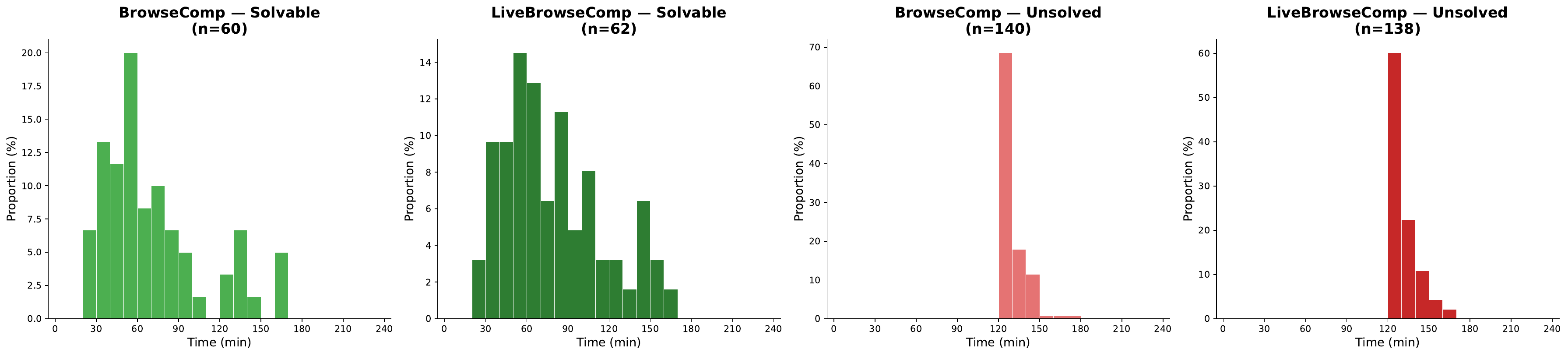}
\caption{Human annotation time distributions on BrowseComp and LiveBrowseComp. Solvers and unsolved questions are shown separately; the proportions and time distributions are closely matched between the two benchmarks.}
\label{fig:human_dist}
\end{figure}

To ensure that LiveBrowseComp and BrowseComp have comparable search difficulty, we screened and trained annotators to strive to achieve this in Stage~4. To further validate this calibration, we recruit a separate group of annotators who did not participate in question construction. Each annotator solves questions from both BrowseComp and LiveBrowseComp using only web search, and must spend at least two hours before giving up on a question. Figure~\ref{fig:human_dist} shows that solve rates are nearly identical (30\% on BrowseComp vs.\ 31\% on LiveBrowseComp), and that completion-time distributions closely match across the two benchmarks. Since human searchers are unaffected by IKD, this result confirms that the two benchmarks are comparable in search difficulty, and that any model-performance drop on LiveBrowseComp (Section~\ref{sec:experiments}) reflects the removal of IKD rather than harder questions.

\vspace{-2mm}
\section{Experimental Evaluation on LiveBrowseComp}
\label{sec:experiments}

The pilot study identifies IKD on static benchmarks; the construction of LiveBrowseComp is designed to neutralize it. We now evaluate model performance on LiveBrowseComp and verify that IKD is effectively suppressed.

\subsection{Experimental Setup}
\label{sec:exp:setup}

We evaluate 11 models spanning a broad range of model families and parameter scales, covering the current frontier of search-capable agents. The open-source group includes DeepSeek-V4-Pro~\cite{deepseek_v4_2026}, Kimi-K2.6~\cite{kimi_k2_6_2026}, Kimi-K2.5~\cite{Bai2026KimiKV}, GLM-5.1~\cite{zeng2026glm}, GLM-5.0~\cite{zeng2026glm}, DeepSeek v3.2~\cite{liu2025deepseek}, and MiniMax-M2.5~\cite{minimax_m2_5_2026}, ranging from 230B to 1.6T parameters. The closed-source group includes Seed-2.0~\cite{seed2026seed2}, GPT-5.4~\cite{openai_gpt54_2026}, Gemini-3.1-Pro~\cite{google_gemini31pro_2026}, and Claude-Sonnet-4.6~\cite{anthropic_sonnet46_2026}, representing the leading commercial API offerings. This selection ensures that our evaluation covers the diversity of current search-agent capabilities rather than reflecting a single model family or deployment paradigm.

All models use the same unified search-agent scaffold as the pilot study~\cite{chu2026redsearcher}, with a shared interaction protocol, sampling budget, context limit, and answer format. Each model uses the system prompt from its official technical report or API documentation, with shared hyperparameters ($\text{temperature}=0.7$, $\text{top\_p}=0.9$). Models are equipped with \textbf{search(query)} (web search via serper.dev, up to 10 results), \textbf{visit(url, goal)} (full page retrieval via Jina with a stated information goal), and \textbf{code\_sandbox} (sandboxed Python interpreter). The maximum context per sample is 256k tokens with a 250-step iteration budget. Many production search agents employ context management strategies such as summarization or retrieval over prior rounds to extend effective context; we do not apply these strategies in our evaluation, which may lower absolute scores but removes a confound. Since all models share the same scaffold, cross-model comparisons remain fair. Full configuration details are provided in Appendix~\ref{app:config}.

\subsection{Main Results}
\label{sec:exp:main}

\begin{table}[t]
\centering
\small
\caption{Model performance across static benchmarks and LiveBrowseComp (avg@4).
LiveBrowseComp produces lower scores and different rankings, reflecting reduced reliance on parametric knowledge.}
\label{tab:main_results}
\begin{tabular*}{\columnwidth}{@{\extracolsep{\fill}} l c c c c c}
\toprule
& & \multicolumn{3}{c}{\textbf{Static Benchmarks}} & \multicolumn{1}{c}{\textbf{Live Benchmark}} \\
\cmidrule(lr){3-5} \cmidrule(lr){6-6}
\textbf{Model} & \textbf{Params} & \textbf{BrowseComp} & \textbf{BrowseComp-ZH} & \textbf{HLE} & \textbf{LiveBrowseComp} \\
\midrule
\multicolumn{6}{l}{\textbf{Open-source models}} \\
\midrule
DeepSeek V4 Pro   & 1.6T & \cellcolor{rank3}61.4 & \cellcolor{rank2}74.6 & \cellcolor{rank1}45.9 & \cellcolor{rank1}38.3 \\
Kimi-K2.6         & 1T   & \cellcolor{rank2}62.4 & \cellcolor{rank1}74.8 & \cellcolor{bluerank3}34.7 & 31.7 \\
Kimi-K2.5         & 1T   & 60.6 & \cellcolor{bluerank1}62.3 & \cellcolor{bluerank2}29.4 & \cellcolor{bluerank3}30.5 \\
GLM 5.1           & 754B & \cellcolor{rank1}68.0 & \cellcolor{rank3}73.5 & \cellcolor{rank2}43.6 & \cellcolor{rank3}33.9 \\
GLM 5.0           & 744B & \cellcolor{bluerank2}59.0 & 68.7 & \cellcolor{rank3}39.5 & \cellcolor{bluerank2}28.5 \\
DeepSeek v3.2     & 671B & \cellcolor{bluerank1}51.4 & \cellcolor{bluerank2}65.0 & 37.1 & \cellcolor{rank2}37.6 \\
MiniMax M2.5      & 230B & \cellcolor{bluerank3}60.4 & \cellcolor{bluerank3}66.1 & \cellcolor{bluerank1}27.1 & \cellcolor{bluerank1}28.0 \\
\midrule
\multicolumn{6}{l}{\textbf{Closed-source models}} \\
\midrule
Seed 2.0          & --- & \cellcolor{rank1}77.3 & \cellcolor{rank1}79.2 & \cellcolor{rank1}54.8 & \cellcolor{rank3}41.5 \\
GPT 5.4           & --- & \cellcolor{rank2}72.1 & \cellcolor{rank3}75.3 & \cellcolor{rank2}51.9 & \cellcolor{rank1}43.2 \\
Gemini 3.1 Pro    & --- & 67.0 & 74.8 & 45.9 & 40.0 \\
Claude Sonnet 4.6 & --- & \cellcolor{rank3}69.3 & \cellcolor{rank2}78.2 & \cellcolor{rank3}49.8 & \cellcolor{rank2}41.4 \\
\bottomrule
\end{tabular*}
\end{table}

Table~\ref{tab:main_results} reports search-augmented performance across static benchmarks and LiveBrowseComp. On LiveBrowseComp, avg@4 ranges from 28.0 (MiniMax M2.5) to 43.2 (GPT 5.4), a sharp drop from the 51--77 point range these models achieve on BrowseComp.

More revealing than the absolute drop is the change in rankings. GLM 5.1 leads BrowseComp at 68.0 but falls to 33.9 on LiveBrowseComp; DeepSeek v3.2, which ranks near the bottom on BrowseComp at 51.4, rises to 37.6 on LiveBrowseComp and outperforms several models that were ahead of it on static benchmarks. This divergence is consistent with unequal IKD across models: models whose static scores were most inflated by parametric knowledge show the largest drops once that advantage is removed.

The inter-model gaps also compress. On BrowseComp, the top-to-bottom spread among open-source models is 16.6 points (68.0 vs.\ 51.4); on LiveBrowseComp it shrinks to 10.3 points (38.3 vs.\ 28.0). IKD amplifies differences between models by rewarding knowledge breadth; once removed, the remaining spread reflects differences in search strategy alone.

\subsection{Closed-Book Validation: IKD Is Effectively Suppressed}
\label{sec:exp:closedbook}

\begin{figure}[t]
    \vspace{-3mm}
  \centering
  \includegraphics[width=\linewidth]{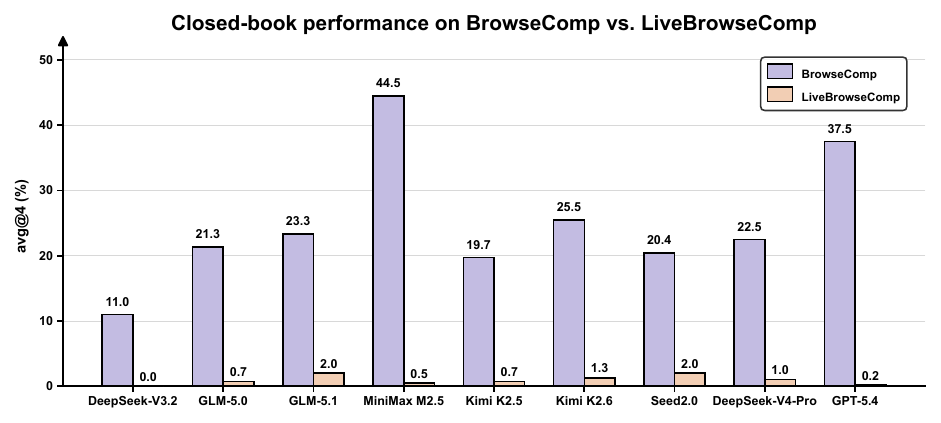}
  \caption{Closed-book performance on BrowseComp-Plus vs.\ LiveBrowseComp. All models fall below 2\% on LiveBrowseComp, confirming that the temporal constraint effectively suppresses parametric knowledge.}
  \label{fig:closedbook_comparison}
\end{figure}

We further verify that the temporal and long-tail constraints indeed suppress IKD. Using the same closed-book configuration as the pilot study (Section~\ref{sec:pilot:closedbook}), we test all models on LiveBrowseComp without any search tools. As shown in Figure~\ref{fig:closedbook_comparison}, every model falls below 2\% closed-book accuracy, compared with 20--44\% on BrowseComp-Plus. Parametric-knowledge coverage is reduced to near zero: on LiveBrowseComp, there is no memory shortcut, and models must search to score.

\subsection{Correlation Analysis: Do Static Benchmarks Predict Live Search?}
\label{sec:exp:correlation}

\begin{figure}[t]
  \centering
  \includegraphics[width=\linewidth]{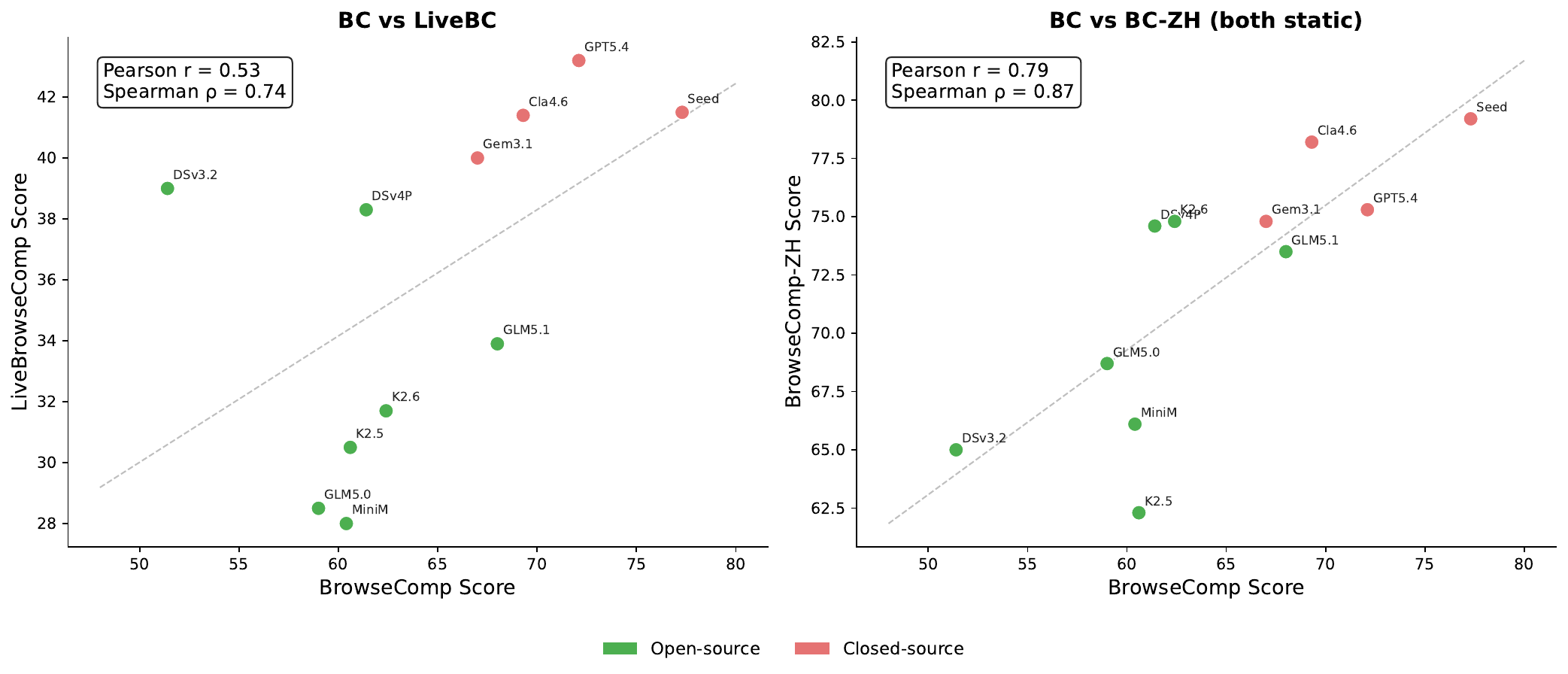}
  \caption{Score correlation between BrowseComp and LiveBrowseComp (left) vs.\ between BrowseComp and BrowseComp-ZH (right). The weaker BC--LiveBC correlation indicates that static benchmark rankings do not reliably transfer to live search evaluation.}
  \label{fig:correlation}
\end{figure}

We examine whether static benchmark rankings transfer to LiveBrowseComp. Figure~\ref{fig:correlation} compares correlations in two settings: BrowseComp vs.\ BrowseComp-ZH (both static), and BrowseComp vs.\ LiveBrowseComp. The Spearman rank correlation drops from $\rho = 0.87$ to $\rho = 0.74$, and the Pearson correlation drops from $r = 0.79$ to $r = 0.53$. The drop is visible in the Pearson coefficient, which is sensitive to absolute score differences: models that benefit most from IKD on static benchmarks do not retain that advantage on LiveBrowseComp. In other words, a model's position on a static leaderboard partly reflects how much it already knows rather than how well it can search, and this knowledge advantage does not transfer to questions outside its training coverage.

\subsection{Turn Distribution Analysis: Search Becomes Exploratory Without IKD}
\label{sec:exp:turns}

\begin{figure}[t]
\vspace{-3mm}
  \centering
  \includegraphics[width=\linewidth]{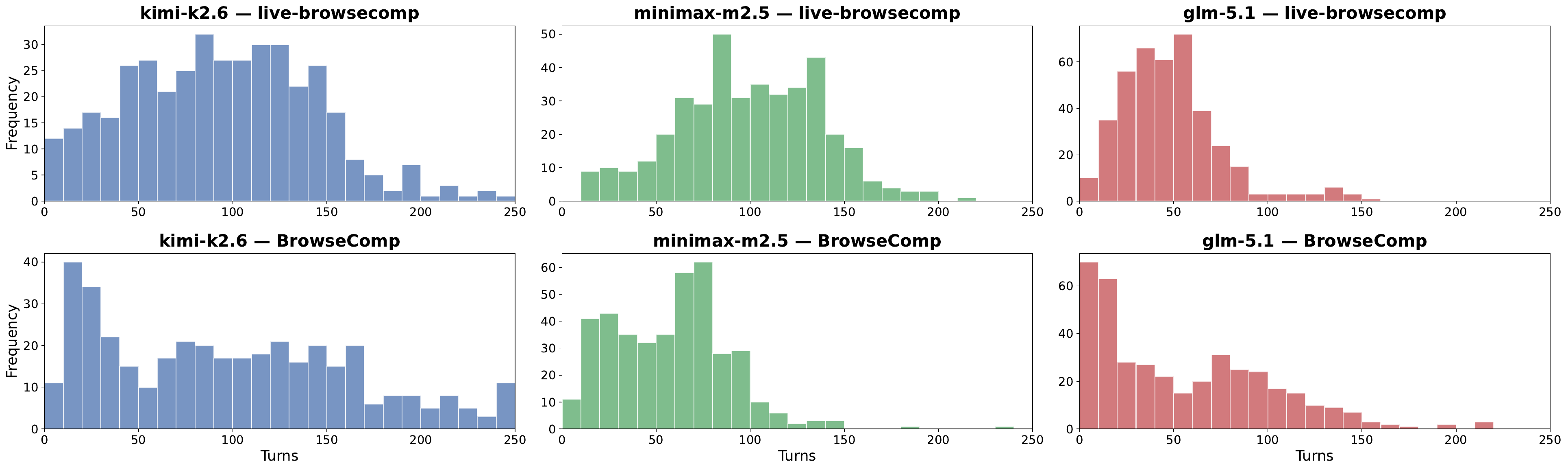}
  \caption{Distribution of search turns per question on BrowseComp vs.\ LiveBrowseComp.}
  \label{fig:turns}
\end{figure}

As further analysis, we examined the search-turn distributions on BrowseComp and LiveBrowseComp (Figure~\ref{fig:turns}). On BrowseComp, a pronounced cluster of questions is resolved within very few turns, consistent with the rapid memory-backed verification pattern identified in Section~\ref{sec:pilot:trajectory}: agents already suspect the answer and use one or two searches to confirm it. On LiveBrowseComp, this short-turn cluster largely disappears, and the distribution shifts toward a single peak at higher turn counts. The implication is behavioral: when agents cannot anchor their search in prior knowledge, each query must actually advance the investigation rather than merely confirm a hypothesis. The resulting trajectories are longer and more exploratory, reflecting genuine information-seeking rather than retrieval-as-verification.

\section{Related Work}
\label{sec:related}

\subsection{Benchmarks for Search Agents}

Evaluation benchmarks for search agents have evolved through several stages. Early retrieval-based QA benchmarks such as NaturalQuestions~\cite{kwiatkowski2019natural}, TriviaQA~\cite{joshi2017triviaqa}, and HotpotQA~\cite{yang2018hotpotqa} focus on fact extraction from static corpora. WebArena~\cite{zhou2023webarena}, Mind2Web~\cite{deng2023mind2web}, and WebVoyager~\cite{he2024webvoyager} extended evaluation to action-level web manipulation. Recently, benchmarks targeting deep search capability have proliferated. BrowseComp~\cite{browsecomp2025} employs an inverted construction methodology with 1,266 questions requiring sustained browsing. GAIA~\cite{mialon2023gaia} designs general-purpose tool-use evaluation tasks. FRAMES~\cite{krishna2024frames} requires multi-step retrieval reasoning across multiple documents. HLE~\cite{phan2025hle} crowdsources challenging academic questions from domain experts worldwide. DeepSearchQA~\cite{gupta2026deepsearchqa} and WideSearch~\cite{wang2025widesearch} target deep reasoning and breadth-oriented search, respectively. Online-Mind2Web~\cite{xue2025illusion} conducts online evaluation on live websites and reveals that web agents tend to exploit shortcuts such as external search engines to bypass intended interaction tasks.

\subsection{Data Contamination and Parametric Knowledge Leakage}

Data contamination has received extensive attention in LLM evaluation. Traditional contamination research~\cite{sainz2023contamination,jacovi2023contamination,deng2024contamination} focuses on literal string overlap between benchmark content and training corpora, proposing detection methods including n-gram overlap, membership inference attacks, temporal cutoff analysis, and behavioral manipulation~\cite{golchin2023dcq}. Beyond literal overlap, recent work has begun to examine related parametric knowledge issues: Du et al.~\cite{du2024context} show that models preferentially rely on internal prior knowledge even when contextual evidence is available, and Ruis et al.~\cite{ruis2024procedural} find that models reason through procedural knowledge synthesis rather than simple retrieval of previously seen answers. Our work addresses a structurally different problem: rather than direct contamination of benchmark data into training corpora, we show that broad parametric knowledge acquired during large-scale pretraining can cover the facts required by benchmark questions, creating a systematic evaluation bias---Intrinsic Knowledge Dependence---that conventional decontamination checks cannot detect.

\subsection{Dynamic and Live Evaluation}

The shift from static to dynamic evaluation is an important recent trend. LiveBench~\cite{white2024livebench} prevents data contamination through monthly refresh cycles and strict temporal cutoffs. LiveCodeBench~\cite{jain2024livecodebench} continuously scrapes new competition problems to guarantee novelty. FreshQA~\cite{vu2023freshqa} maintains temporally-aware factoid questions with frequency-graded update tiers. LiveMathBench~\cite{liu2025livemathbench} employs periodic difficulty-split versions to test frontier mathematical reasoning. The Self-Evolving Benchmark framework~\cite{loong2025selfevolving} argues for dynamic instance reframing to combat shortcut biases.

\section{Discussion and Conclusion}
\label{sec:discussion}

This paper identifies \emph{Intrinsic Knowledge Dependence} as a central confound in search-agent evaluation: agents can score well by generating hypotheses from parametric knowledge and using search mainly for confirmation. 
LiveBrowseComp addresses this by using recent, long-tail questions that place agents beyond their current knowledge coverage.
On this benchmark, closed-book accuracy falls below 2\%, search-augmented scores drop sharply, and model rankings change, while human search effort remains comparable to BrowseComp. 
These findings argue for dynamic, time-sensitive benchmarks as a standard part of search-agent evaluation, and for training signals that reward evidence-led discovery rather than guess-and-verify behavior.


\bibliographystyle{unsrtnat}
\bibliography{references}

\appendix

\section{Data Source and Filtering Criteria}
\label{app:datasources}

Table~\ref{tab:data_sources} provides detailed specifications of the API interfaces, filtering thresholds, and long-tail selection criteria for each data source.

\begin{table}[h]
\centering
\caption{Data source specifications and filtering criteria.}
\label{tab:data_sources}
\small
\begin{tabular}{lllll}
\toprule
\textbf{Source} & \textbf{API} & \textbf{Window} & \textbf{Filter Metric} & \textbf{Threshold} \\
\midrule
GDELT & GDELT API & 90 days & LLM heat score, article length & 2.0--4.0, $>$150 chars \\
TMDB & TheMovieDB API & 90 days & Popularity, vote\_count, revenue & $\ge 2.5$ long-tail \\\\
RAWG & RAWG API & 90 days & Ratings\_count, added, Metacritic & $\ge 2.5$ long-tail \\\\
CVE & CVE/NVD API & 90 days & CVSS, recency, exploit, refs & $\ge 2.0$ long-tail \\\\
Sports & Match data API & 90 days & League tier, attendance, diversity & $\ge 1.5$ long-tail \\\\
USGS & USGS Earthquake API & 90 days & Magnitude, significance, depth & $\ge 1.5$ long-tail \\\\
\bottomrule
\end{tabular}
\end{table}

\paragraph{GDELT (Global Events).} GDELT's raw news stream is the noisiest source (459k articles filtered down to 1k). An LLM scorer rates each article on three dimensions: geographic audience, entity fame, and societal impact. Articles with average scores in the 2.0--4.0 range are retained---below this threshold are mostly spam or junk pages, above it are global headlines whose answers would be too easily guessable. Hard filters additionally discard paywalls, bot-detection pages, 404 errors, and articles under 150 characters.

\paragraph{TMDB (Film \& Television).} Filtering is entirely heuristic. The long-tail score rewards low popularity ($\leq 1$ maximum points), low vote count ($\leq 20$ maximum points), and zero box-office revenue (maximum points, indicating the film likely had no theatrical release). Non-English-language films receive a bonus. Entries must also have an overview, at least three cast members, and an IMDB or Wikidata identifier. Only items with total long-tail score $\geq 2.5$ are retained.

\paragraph{RAWG (Games).} Similar to TMDB but with game-specific dimensions. Obscurity is measured by ratings\_count ($\leq 10$ maximum points) and ``added'' count ($\leq 1$ maximum points). A unique dimension rewards the \emph{absence} of a Metacritic score (maximum points), while scores $\geq 85$ receive zero. Non-ASCII game names receive a bonus. Entries must have at least one developer or publisher and a non-empty description; total long-tail score $\geq 2.5$ is required.

\paragraph{CVE (Cybersecurity).} The CVE database is relatively small (approximately 1,500 entries). Scoring dimensions include severity (CVSS $\geq 9.0$ maximum points), specificity (vulnerabilities affecting a single product score highest), exploit availability, recency (within 30 days maximum points), and reference richness. Only entries with English descriptions, non-REJECTED status, and valid CVSS scores are retained. Total long-tail score $\geq 2.0$ suffices.

\paragraph{Sports (Athletic Events).} Two signal classes are used: match completeness (must have scores, result descriptions, and finished status) and significance (keyword extraction from event names for terms like ``final'' or ``championship,'' attendance above 50,000). A diversity bonus of $+1.5$ is awarded to non-football events, since football dominates the raw data and would produce overly predictable questions. Total long-tail score $\geq 1.5$ is required.

\paragraph{USGS (Earthquake Data).} The filter targets events that are ``felt but not catastrophic.'' Magnitude is the primary dimension (M7+ maximum points, M3--M4 near minimum). Significance (USGS composite indicator $\geq 600$ maximum), depth (very shallow $<10$ km or very deep $>500$ km score high), and impact indicators (tsunami alerts, felt reports, PAGER color) are also considered. Events must have location annotations. Total long-tail score $\geq 1.5$ suffices.

\section{Scoring Prompt}
\label{app:scoring}

We use an LLM-as-judge evaluator (GPT-OSS) to score model outputs against reference answers. The judge extracts a final answer from the model's response and compares it to the ground-truth answer, allowing reasonable surface-form variants (abbreviations, aliases) while rejecting imprecise or partially correct answers. The prompt template is shown below.

\begin{tcolorbox}[
  title=LLM-as-Judge Prompt,
  colback=blue!2!white,
  colframe=blue!30!black,
  fonttitle=\bfseries\small,
  fontupper=\ttfamily\footnotesize,
  left=2mm, right=2mm, top=1mm, bottom=1mm
]
Judge whether the following [response] to [question] is correct or not based on the precise and unambiguous [correct\_answer] below.\\[2pt]
[question]: \{question\}\\[2pt]
[response]: \{response\}\\[2pt]
Your judgement must be in the format and criteria specified below:\\[2pt]
extracted\_final\_answer: The final exact answer extracted from the [response]. Put the extracted answer as `None' if there is no exact, final answer to extract from the response.\\[2pt]
[correct\_answer]: \{correct\_answer\}\\[2pt]
reasoning: Explain why the extracted\_final\_answer is correct or incorrect based on [correct\_answer], focusing only on if there are meaningful differences between [correct\_answer] and the extracted\_final\_answer. Do not comment on any background to the problem, do not attempt to solve the problem, do not argue for any answer different than [correct\_answer], focus only on whether the answers match.\\[2pt]
correct: Answer `yes' if extracted\_final\_answer matches the [correct\_answer] given above, or is within a small margin of error for numerical problems. Answer `no' otherwise, i.e.\ if there is any inconsistency, ambiguity, non-equivalency, or if the extracted answer is incorrect.\\[2pt]
confidence: The extracted confidence score between 0\% and 100\% from [response]. Put 100 if there is no confidence score available.
\end{tcolorbox}

\medskip
If the judge's output begins with ``A'', it is treated as correct; if it begins with ``B'', it is treated as incorrect. Failed parses are retried up to five times; after five failures the answer is marked incorrect.

\section{Search Agent Experimental Configuration}
\label{app:config}

\subsection{General Configuration}
\label{app:config:general}

\paragraph{System prompt.} The default system prompt used for most models (DeepSeek, GLM, MiniMax, Seed) follows the deep search assistant template below. Kimi uses its own cookbook-aligned prompt, and Seed aligns with the Seed1.8 Cookbook format.

\begin{tcolorbox}[
  title=System Prompt (Default),
  colback=green!2!white,
  colframe=green!30!black,
  fonttitle=\bfseries\small,
  fontupper=\ttfamily\footnotesize,
  left=2mm, right=2mm, top=1mm, bottom=1mm
]
You are a deep search assistant. Your primary role is to perform rigorous, multi-step, multi-source investigations on any topic---covering both broad, open-domain questions and highly specialized academic inquiries. For each user request, you must actively seek out and cross-check information from credible and diverse sources, then integrate the findings into a response that is comprehensive, accurate, well-structured, and objective.\\[2pt]
Operating principles:\\
1. Plan and execute research: Break complex questions into sub-questions, gather evidence across multiple sources, and prioritize primary sources and authoritative references when available.\\
2. Evaluate source quality: Prefer reputable institutions, peer-reviewed research, official documentation, and high-quality journalism. Note uncertainty, conflicts, and limitations when sources disagree.\\
3. Synthesize, don't just list: Combine evidence into a coherent narrative or structured output, highlighting key takeaways and nuanced trade-offs.\\
4. Maintain neutrality: Present competing viewpoints fairly when relevant, and avoid unsupported speculation.\\[2pt]
When you have collected sufficient information and are ready to deliver the definitive response, you must wrap the entire final answer in $<$answer$><$/answer$>$ tags.
\end{tcolorbox}

\paragraph{Tools.} In the LiveBrowseComp experiments, models have access to \texttt{search(query)} (web search via serper.dev, returning up to 10 results with URLs and text snippets), \texttt{visit(url, goal)} (open and retrieve full page content summarized toward an information goal), and, depending on the model, a Python interpreter, Google Scholar, and Google Maps.

\paragraph{Context limit and forced answer.} The maximum context length per sample is 256k tokens. When a model exceeds this limit or the maximum iteration budget (250 steps), a forced-answer prompt is injected to elicit a final response:

\begin{tcolorbox}[
  title=Force-Answer Prompt,
  colback=orange!2!white,
  colframe=orange!30!black,
  fonttitle=\bfseries\small,
  fontupper=\ttfamily\footnotesize,
  left=2mm, right=2mm, top=1mm, bottom=1mm
]
You have reached the maximum number of research steps or context limit. Based on all the information you have gathered so far, please provide your final answer to the original question. Put your final answer between $<$answer$>$ and $<$/answer$>$.
\end{tcolorbox}

\subsection{BrowseComp-Plus and Dense Retrieval Experiments}
\label{app:config:bcplus}

BrowseComp-Plus is a curated subset of BrowseComp, expanded with a comprehensive document library provided by the benchmark's creators. For each question, this library contains four categories of documents: \emph{evidence documents} (containing direct evidence for the answer), \emph{gold documents} (high-quality supporting material), \emph{irrelevant documents} (distractors unrelated to the question), and \emph{hard-negative documents} (superficially relevant but ultimately unhelpful). This annotation enables precise tracing of whether a model makes substantive contact with the correct answer during search, allowing fine-grained behavioral analysis.

For the pilot study in Section~\ref{sec:pilot}, we follow the official BrowseComp-Plus recommendations and construct a dense retrieval index over the provided document library using the Qwen3-8B-Embedding model, which serves as the model's search environment. For the evidence-blocking experiments (\S\ref{sec:pilot:drop}), we intentionally remove all evidence documents and gold documents when building the index, retaining only irrelevant and hard-negative documents. Under this condition, models can still issue searches but can never retrieve information that supports the correct answer. In all dense retrieval experiments, Google Scholar and Google Maps are disabled, and models are prohibited from accessing the internet within the Python interpreter.

\section{Limitations}

LiveBrowseComp has several limitations. First, the 90-day temporal window is an approximate heuristic: some facts produced within the window may have been announced or leaked earlier, and different models may have different training cutoffs, so the boundary is not perfectly crisp. Second, all evaluations use a single search backend (serper.dev); different search indices may yield different results, and a model's apparent search capability may partly reflect the coverage of the underlying index rather than its own search strategy. Third, the reliance on expert human annotation and multi-stage verification, while essential for quality, makes the benchmark difficult to scale and incurs a high per-question cost.

\section{Closed-Book Answering Configuration}
\label{app:closedbook}

In the closed-book experiments of Sections~\ref{sec:pilot:closedbook}, models answer questions using only parametric knowledge, without any search or browsing tools. All models share a unified system prompt:

\begin{tcolorbox}[
  title=Closed-Book System Prompt,
  colback=blue!2!white,
  colframe=blue!30!black,
  fonttitle=\bfseries\small,
  fontupper=\ttfamily\footnotesize,
  left=2mm, right=2mm, top=1mm, bottom=1mm
]
You are an expert knowledge assistant. Answer the question below using your own knowledge only.\\[2pt]
IMPORTANT RULES:\\
- Break down each clue, cross-check against your knowledge, and commit to your best answer.\\
- Do NOT overthink or repeat yourself. Once you have a candidate that fits most clues, commit to it.\\
- When uncertain, make your best educated guess rather than analyzing endlessly.\\
- You MUST wrap your final answer in $<$answer$><$/answer$>$ tags.
\end{tcolorbox}

Due to architectural differences across model APIs, the implementation falls into three categories. \textbf{Single-round chat completion} (DeepSeek V4, MiniMax M2.5): the answer is obtained in one API call with thinking disabled (DeepSeek) or uncontrolled (MiniMax), using max\_tokens of 16384 and 8192 respectively. \textbf{Two-phase with prefill} (GLM-5, Kimi K2.5/K2.6, MiroThinker): Phase~1 enables thinking for deep reasoning; if the reasoning length exceeds the context budget and the model produces no content output (finish\_reason=length), the thinking is truncated and Phase~2 uses the completions endpoint to prefill the truncated reasoning, guiding the model to produce the final answer. The prefill strategy varies by model: GLM-5 directly prefills to \texttt{$<$/think$>$\textbackslash n$<$answer$>$}, while Kimi first prefills only the closing think tag and falls back to prefill with \texttt{$<$answer$>$} if the model still does not produce an answer. \textbf{Reasoning-effort control} (Seed 2.0 Pro): a single call with reasoning\_effort=high, placing reasoning in a dedicated reasoning\_content field without a two-phase pipeline.

All closed-book experiments use 4 independent samples, matching the tool-use experiments, and report both pass@4 and avg@4.

\section{Human Annotation Details}
\label{app:annotation}

\subsection{Annotation Workflow and Quality Control}

The construction of LiveBrowseComp involves two distinct roles: \emph{annotators}, who draft questions and reference answers from filtered seed events, and \emph{verifiers}, who independently validate each question. All annotators have passed the screening and training process described in the main text.

\paragraph{Stage 4: Question construction.} Each annotator receives a seed event with its source URL and basic metadata. Annotation guidelines specify the three construction criteria: (a)~multi-step, multi-source difficulty comparable to the BrowseComp questions solved during screening, where the answer cannot be found in the first three pages of search results for the question text; (b)~verifiability from definitive sources with a single short-string answer; (c)~temporal anchoring requiring at least one clue from the past 90 days. An example seed event is shown below.

\begin{tcolorbox}[
  title=Annotator Input: Seed Event (Example),
  colback=blue!2!white,
  colframe=blue!30!black,
  fonttitle=\bfseries\small,
  fontupper=\ttfamily\footnotesize,
  left=2mm, right=2mm, top=1mm, bottom=1mm
]
Source: TMDB\\
URL: https://www.themoviedb.org/movie/1641989\\
Title: Where do I go from Here? (2026)\\
Release Date: 2026-03-02 (GB)\\
Genres: Documentary, Drama\\
Runtime: 8m
\end{tcolorbox}

Annotators independently conduct web research starting from the seed event and craft a question that meets the three criteria above. They record all web pages they visit and the evidence chain linking the question to the answer. The example below shows an annotator's output for the seed event above.

\begin{tcolorbox}[
  title=Annotator Output (Example),
  colback=orange!2!white,
  colframe=orange!30!black,
  fonttitle=\bfseries\small,
  fontupper=\ttfamily\footnotesize,
  left=2mm, right=2mm, top=1mm, bottom=1mm
]
\textbf{Question:} Imagine a film project created as a collaboration between two production companies: one named after a rigorous financial planning method from the 1970s, where each budget cycle starts from zero; the other known by a five-letter name representing independent creative projects in film or music. This art film explores profound themes of diaspora and the elusive emotional search for ``home,'' with its narrative and performances all handled by a single artist who is both director and lead actor. What is the title of this film?\\[4pt]
\textbf{Answer:} Where do I go from Here?\\[4pt]
\textbf{Evidence chain (excerpt):}\\
\quad 1. https://www.themoviedb.org/movie/1641989 \textnormal{(seed: film metadata, release date, genres)}\\
\quad 2. https://www.zerobased.co.uk/about \textnormal{(production company: ``zero-based budgeting'' origin)}\\
\quad 3. https://www.indiefilmpage.com/where-do-i-go \textnormal{(second production company, diaspora theme, single-artist credit)}\\
\quad \ldots
\end{tcolorbox}

This documentation serves as the primary input for the subsequent verification stages.

\paragraph{Stage 5(a): Correctness and uniqueness.}
Each verifier receives the annotator's question, reference answer, and the full evidence chain.

\paragraph{Uniqueness protocol.}
To generate a broad pool of candidate answers, we use DeepSeek-V4-Pro, GLM-5.1, Kimi-K2.6, and MiniMax-M2.5, each rolled out 8 times per question, with search tools enabled. All generated candidates, regardless of whether they match the reference answer, are collected. Verifiers then manually inspect each candidate and check whether it satisfies every constraint stated in the question. If any alternative answer passes all constraints, the question is removed. This protocol is deliberately conservative: it may eliminate some genuinely valid questions, but it maximizes the likelihood that every retained question has a unique answer, even though absolute uniqueness cannot be guaranteed.

The following shows the verifier's task sheet for the running example.

\begin{tcolorbox}[
  title=Verifier Task: Correctness and Uniqueness (Example),
  colback=blue!2!white,
  colframe=blue!30!black,
  fonttitle=\bfseries\small,
  fontupper=\ttfamily\footnotesize,
  left=2mm, right=2mm, top=1mm, bottom=1mm
]
\textbf{Question:} Imagine a film project created as a collaboration between two production companies: one named after a rigorous financial planning method from the 1970s, where each budget cycle starts from zero; the other known by a five-letter name representing independent creative projects in film or music. This art film explores profound themes of diaspora and the elusive emotional search for ``home,'' with its narrative and performances all handled by a single artist who is both director and lead actor. What is the title of this film?\\[2pt]
\textbf{Reference answer:} Where do I go from Here?\\[2pt]
\textbf{Evidence chain:}\\
\quad URL 1: https://www.themoviedb.org/movie/1641989\\
\quad\quad -- Film metadata, release date 2026-03-02, genres: Documentary, Drama\\
\quad URL 2: https://www.zerobased.co.uk/about\\
\quad\quad -- Production company named after zero-based budgeting (1970s financial method)\\
\quad URL 3: https://www.indiefilmpage.com/where-do-i-go\\
\quad\quad -- Second production company (5-letter name), diaspora theme, single director-actor\\
\quad \ldots\\[2pt]
\textbf{Task (Part A -- Correctness):} Visit each URL. Confirm that each cited page supports the claimed fact, and that the logical chain from clues to answer is valid.\\[2pt]
\textbf{Task (Part B -- Uniqueness):} The following LLM-generated candidate answers have been collected:\\
\quad 1. ``Where do I go from Here?'' \textnormal{(Model A, Model C, Model E -- matches reference)}\\
\quad 2. ``The Diaspora Project'' \textnormal{(Model B)}\\
\quad 3. ``Zero Based'' \textnormal{(Model D)}\\
\quad 4. ``Homecoming'' \textnormal{(Model F)}\\
\quad \ldots\\
For each candidate $\neq$ reference answer, manually search the web and check whether it satisfies \emph{every} constraint in the question.\\[2pt]
\textbf{Result:} PASS / FAIL \textnormal{(if FAIL, specify broken evidence or alternative valid answer)}\\[2pt]
\textnormal{(Three verifiers independently complete this task per question.)}
\end{tcolorbox}

\paragraph{Stage 5(b): Difficulty screening.}
Three independent annotators who were not involved in question construction or Stage~5(a) attempt to solve each question using only web search.

\begin{tcolorbox}[
  title=Annotator Task: Difficulty Screening (Example),
  colback=blue!2!white,
  colframe=blue!30!black,
  fonttitle=\bfseries\small,
  fontupper=\ttfamily\footnotesize,
  left=2mm, right=2mm, top=1mm, bottom=1mm
]
\textbf{Question:} \textnormal{(same as above)}\\[2pt]
\textbf{Task:} Solve the question using only web search. Record all pages visited and the evidence chain. Note the total time spent.\\[2pt]
\textbf{Result:} SOLVED within 30 min / NOT SOLVED within 30 min \textnormal{(if SOLVED, the question is excluded)}\\[2pt]
\textnormal{(Three annotators independently complete this task per question.)}
\end{tcolorbox}

\paragraph{Stage 5(c): Temporality verification.}
Each verifier receives the question, the reference answer, and the evidence chain with publication dates annotated.

\begin{tcolorbox}[
  title=Verifier Task: Temporality (Example),
  colback=blue!2!white,
  colframe=blue!30!black,
  fonttitle=\bfseries\small,
  fontupper=\ttfamily\footnotesize,
  left=2mm, right=2mm, top=1mm, bottom=1mm
]
\textbf{Question:} \textnormal{(same as above)}\\[2pt]
\textbf{Reference answer:} Where do I go from Here?\\[2pt]
\textbf{Evidence chain (with dates):}\\
\quad URL 1: https://www.themoviedb.org/movie/1641989 \texttt{[RECENT -- 2026-02]}\\
\quad URL 2: https://www.zerobased.co.uk/about \texttt{[PRE-WINDOW -- 2020]}\\
\quad URL 3: https://www.indiefilmpage.com/where-do-i-go \texttt{[RECENT -- 2026-03]}\\
\quad \ldots\\[2pt]
\textbf{Task:} For each \texttt{[RECENT]} URL, restrict web search to results published before the 90-day cutoff and attempt to find substitute evidence:\\
\quad $\square$ URL 1 (film metadata, 2026-02): Can you find this film's title, release date, and genre information from before 2026-01? $\rightarrow$ search \ldots\\
\quad $\square$ URL 3 (production details, 2026-03): Can you find the diaspora theme and single director-actor credit from before 2026-01? $\rightarrow$ search \ldots\\[2pt]
\textnormal{URL 2 is \texttt{[PRE-WINDOW]} and does not need substitution.}\\[2pt]
\textbf{Result:} TEMPORAL / NOT-TEMPORAL \textnormal{(if NOT-TEMPORAL, list the substitute URLs found for each \texttt{[RECENT]} page)}\\[2pt]
\textnormal{(Three verifiers independently complete this task per question.)}
\end{tcolorbox}

\paragraph{Quality control.} The three-person independent verification described above applies to \emph{every} check---Stage~5(a), 5(b), and 5(c)---for every question. After all three verifiers complete their task sheets for a given stage, a fourth independent verifier receives the three completed sheets and cross-checks them. The following shows an example cross-check for Stage~5(a).

\begin{tcolorbox}[
  title={Fourth Verifier Task: Cross-Check (Example, Stage 5(a))},
  colback=blue!2!white,
  colframe=blue!30!black,
  fonttitle=\bfseries\small,
  fontupper=\ttfamily\footnotesize,
  left=2mm, right=2mm, top=1mm, bottom=1mm
]
\textbf{Verifier A --- Completed Task Sheet (excerpt):}\\
\quad Result: PASS\\
\quad Notes: All 3 cited pages accessible. Evidence from URL 2 confirms zero-based budgeting as the 1970s method. URL 3 confirms single director-actor and diaspora theme. Logic chain intact.\\[2pt]
\textbf{Verifier B --- Completed Task Sheet (excerpt):}\\
\quad Result: PASS\\
\quad Notes: Evidence chain valid. All constraints supported.\\[2pt]
\textbf{Verifier C --- Completed Task Sheet (excerpt):}\\
\quad Result: FAIL\\
\quad Notes: URL 2 (zerobased.co.uk/about) returns a 404 error. Cannot confirm the claimed fact about zero-based budgeting.\\[2pt]
\textbf{Task:} Review the three completed sheets. Verifiers A and B passed, Verifier C failed due to a broken link. Independently check the disputed point: revisit URL 2. If unreachable, confirm whether the claimed fact can be verified from an alternative source. Issue a final ruling.\\[2pt]
\textbf{Result:} FINAL PASS / FINAL FAIL \textnormal{(with reasoning)}
\end{tcolorbox}

The same cross-check procedure is applied after Step~5(b) and Step~5(c), with the fourth verifier examining the three completed task sheets from each stage and resolving any disagreements.

\subsection{Annotator Compensation}

All annotators and verifiers are employed and compensated by a third-party annotation company at an hourly rate of approximately \$9.60 USD.

\section{Per-Domain Analysis}
\label{app:domain}

Table~\ref{tab:domain} reports model accuracy across the five topical categories of LiveBrowseComp. Considerable performance variation is visible both across domains and across models. Sports and Society~\&~Culture tend to yield higher scores for several models, while Sci.~\&~Tech.\ is more challenging for most. Within-domain rankings also diverge from aggregate rankings: for instance, GLM~5.0 leads Entertainment by a wide margin (52.1\%) despite a below-average overall score (28.5\%), and Kimi~K2.5 achieves the single best domain result at 66.7\% in Sports. Closed-source models generally dominate Movies and Sci.~\&~Tech., but open-source models are competitive in Entertainment and Sports. These patterns suggest that domain-specific knowledge coverage varies across model families, and that aggregate scores alone can mask meaningful capability differences.

\begin{table}[h!]
\centering
\small
\caption{Per-domain accuracy on LiveBrowseComp (avg@4).}
\label{tab:domain}
\begin{tabular}{lcccccc}
\toprule
\textbf{Model} & \textbf{Movies} & \textbf{Entertainment} & \textbf{Sci.\ \& Tech.} & \textbf{Sports} & \textbf{Society\ \& Culture} & \textbf{Avg.} \\
\midrule
DeepSeek v3.2     & 32.5 & 35.0 & 27.5 & 45.0 & 42.0 & 37.6 \\
GLM 5.0           & 30.1 & \cellcolor{rank1}52.1 & 22.2 & 35.0 & \cellcolor{rank2}43.8 & 28.5 \\
GLM 5.1           & 34.3 & 33.5 & 18.5 & \cellcolor{rank2}58.5 & 34.5 & 33.9 \\
MiniMax M2.5      & 16.9 & 20.5 & 23.1 & 37.5 & 34.4 & 28.0 \\
Kimi K2.5         & 25.0 & 21.8 & 30.6 & \cellcolor{rank1}66.7 & \cellcolor{rank3}43.2 & 30.5 \\
Kimi K2.6         & 26.7 & 27.8 & 19.4 & 33.3 & \cellcolor{rank1}52.4 & 31.7 \\
DeepSeek V4 Pro   & 38.7 & 41.1 & \cellcolor{rank3}30.8 & \cellcolor{rank3}50.0 & 40.6 & 38.3 \\
Seed 2.0          & \cellcolor{rank1}51.6 & 42.9 & 30.8 & \cellcolor{rank3}50.0 & 37.5 & 41.5 \\
GPT 5.4           & \cellcolor{rank2}48.0 & \cellcolor{rank2}43.0 & \cellcolor{rank1}40.0 & 46.0 & 39.0 & \cellcolor{rank1}43.2 \\
Gemini 3.1 Pro    & 41.9 & 39.3 & \cellcolor{rank2}38.5 & \cellcolor{rank3}50.0 & 31.2 & 40.0 \\
Claude Sonnet 4.6 & \cellcolor{rank3}45.0 & \cellcolor{rank2}43.0 & \cellcolor{rank3}38.0 & 46.0 & 35.0 & \cellcolor{rank2}41.4 \\
\bottomrule
\end{tabular}
\end{table}

\end{document}